\documentclass[a4paper,twoside]{article}
\usepackage[utf8]{inputenc}
\usepackage[T1]{fontenc}
\usepackage{rfia2000}
\usepackage[english]{babel}
\usepackage{graphicx}
\usepackage{siunitx}
\usepackage{hyperref}
\usepackage{fancyvrb}
\usepackage[capitalize]{cleveref}
\usepackage{booktabs}
\usepackage{subcaption}
\usepackage[subtle]{savetrees}
\usepackage{tabularx}
\newcolumntype{Y}{>{\centering\arraybackslash}X}
\usepackage{xspace}
\usepackage{times}
\usepackage{enumitem}
\setlist{noitemsep}
\setlist{nosep}

\def\vis{\textsc{\textbf{Image}}\xspace} 
\def\tex{\textsc{\textbf{Text}}\xspace} 
\def\fus{\textsc{\textbf{Fusion}}\xspace} 

\begin{document}
\date{}
\title{\Large\bf Multimodal deep networks for text and image-based document classification}
\author{\begin{tabular}[t]{c@{\extracolsep{4em}}c@{\extracolsep{4em}}c@{\extracolsep{4em}}c}
  Nicolas Audebert & Catherine Herold & Kuider Slimani & Cédric Vidal\\
\end{tabular}
{} \\
\\
Quicksign, 38 rue du Sentier, 75002 Paris
  {} \\
\\
\{nicolas.audebert,catherine.herold,kuider.slimani,cedric.vidal\}@quicksign.com\\
}
\maketitle
\thispagestyle{empty}
\subsection*{R\'esum\'e}
{\em
  La classification automatique de documents numérisés est importante pour la dématérialisation de documents historiques comme de procédures administratives.
  De premières approches ont été suggérées en appliquant des réseaux convolutifs aux images de documents en exploitant leur aspect visuel.
  Toutefois, la précision des classes demandée dans un contexte réel dépend souvent de l'information réellement contenue dans le texte, et pas seulement dans l'image.
  Nous introduisons un réseau de neurones multimodal capable d'apprendre à partir d'un plongement lexical du texte extrait par reconnaissance de caractères et des caractéristiques visuelles de l'image.
  Nous démontrons la pertinence de cette approche sur Tobacco3482 et RVL-CDIP, augmentés de notre jeu de données textuel QS-OCR\footnotemark{}, sur lesquels nous améliorons les performances d'un modèle image de 3\% grâce à l'information sémantique textuelle.
}
\subsection*{Mots-clés}
Classification de documents, apprentissage multimodal, fusion de données.

\subsection*{Abstract}
{\em
  Classification of document images is a critical step for archival of old manuscripts, online subscription and administrative procedures.
  Computer vision and deep learning have been suggested as a first solution to classify documents based on their visual appearance.
  However, achieving the fine-grained classification that is required in real-world setting cannot be achieved by visual analysis alone.
  Often, the relevant information is in the actual text content of the document.
  We design a multimodal neural network that is able to learn from word embeddings, computed on text extracted by OCR, and from the image.
  We show that this approach boosts pure image accuracy by 3\% on Tobacco3482 and RVL-CDIP augmented by our new QS-OCR text dataset\footnotemark[\value{footnote}], even without clean text information.
}
\subsection*{Keywords}
Document classification, multimodal learning, data fusion.

\footnotetext{\url{https://github.com/Quicksign/ocrized-text-dataset}}

\setlength\textfloatsep{5pt}
\setlength\dbltextfloatsep{5pt}

\section{Introduction}
\begin{figure}[t]
  \includegraphics[width=\linewidth]{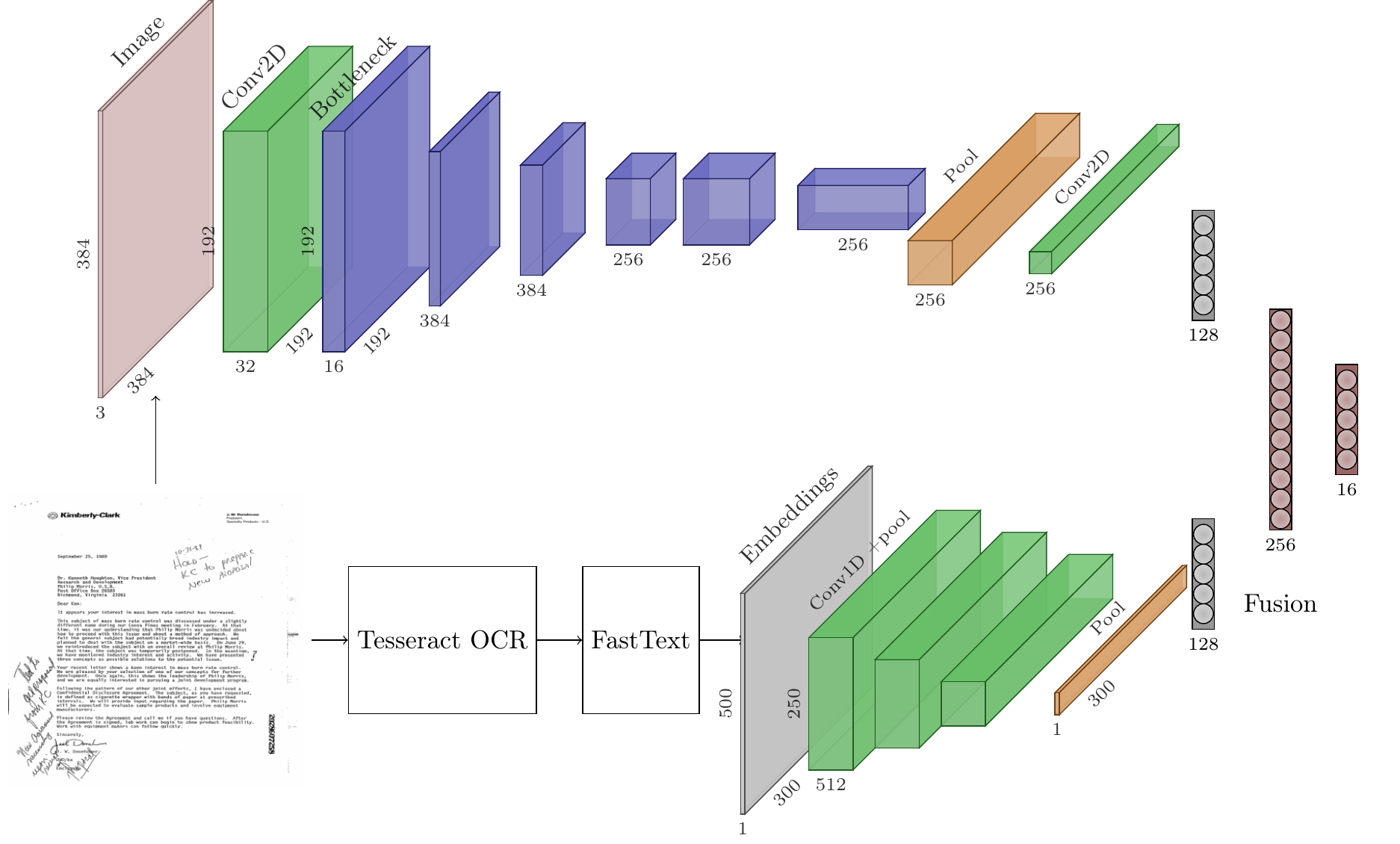}
  \caption{Multimodal classifier for hybrid text/image classification. Training is performed end-to-end on both textual and visual features.}
  \label{fig:multimodal_network}
\end{figure}

\begin{figure*}[t]
  \begin{subfigure}[b]{0.25\textwidth}
    \includegraphics[width=\textwidth]{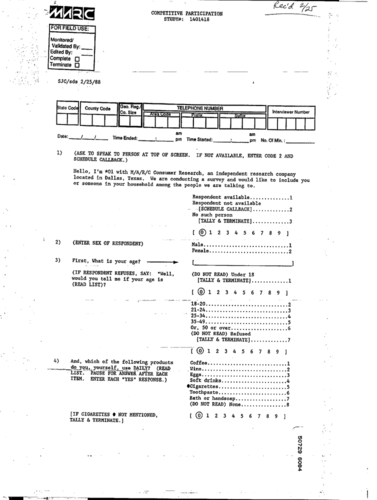}
    \begin{Verbatim}[xleftmargin=1mm,fontsize=\tiny]
Keck Zr COMPETITIVE
PARTICIPATION
sTupYe: 1401618 Sic/sds
2/25/88 .
= Coren aE a Dale:__/{__j__
Time Ended: 4 ‘pm
Time Started:. pm
No. Of Min. :, 1)
(ASK To SPEAK TO PERSON AT
TOP OF SCREEN, IF NOT
AVAILABLE, ENTER CODE 2 AY
SCHEDULE CALLBACK
...

    \end{Verbatim}
    \caption{Questionnaire}
  \end{subfigure}%
  \begin{subfigure}[b]{0.25\textwidth}
    \includegraphics[width=\textwidth]{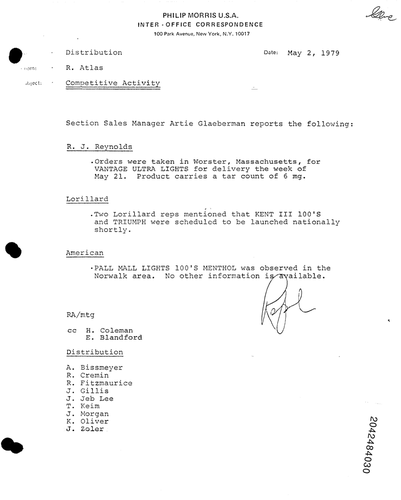}
    \begin{Verbatim}[xleftmargin=1mm,fontsize=\tiny]
PHILIP MORRIS U.S.A. koe
INTER - OFFICE
CORRESPONDENCE
100 Park Avenua, New York,
N.Y. 10017
Distribution Date:
May 2, 1979
R, Atlas
Competitive Activity

Section Sales Manager Artie
Glacberman reports the
following:
...

    \end{Verbatim}
    \caption{Memo}
  \end{subfigure}%
  \begin{subfigure}[b]{0.25\textwidth}
    \includegraphics[width=\textwidth]{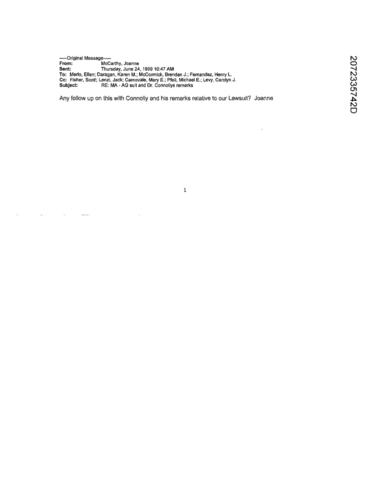}
    \begin{Verbatim}[xleftmargin=1mm,fontsize=\tiny]
Orginal Message-—
From: McCarthy, Josnne
Sent. ‘Thursday, June 24,
1990 10:47 AM
Tor Met, Eten; Daragan,
Karon M; McCormick, Brendan;
Femandez, Henry L.
(Ge: Fisher, Scot: Lon
Jock: Gamavalo, Mary ;
Pol, Michael. Ley, Ceo} J
Subject: RE: MA™-AG out
and Dr. Connolys remarks
‘Any follow up on this with
Connolly and his remarks
relative to our Lawsuit?
Joanne Oztrlseezloz
    \end{Verbatim}
    \caption{Email}
  \end{subfigure}%
  \begin{subfigure}[b]{0.25\textwidth}
    \includegraphics[width=\textwidth]{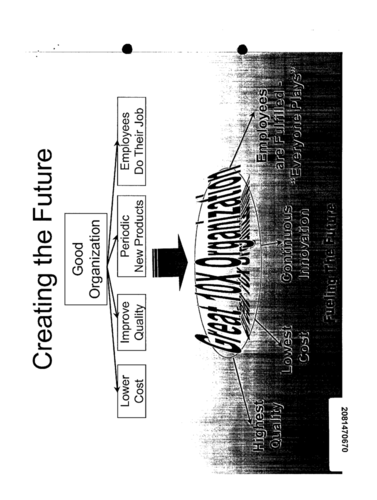}
    \begin{Verbatim}[xleftmargin=1mm,fontsize=\tiny]
    
    
    
Creating the Future
Good
Organization

Lower Improve Periodic
Employees e
Cost Quality New Products
Do Their Job

029021807



    \end{Verbatim}
    \caption{Presentation}
  \end{subfigure}%
  \vspace{-0.5em}
  \caption{Document samples from the RVL-CDIP~\cite{harley_2015_evaluation} dataset with corresponding text extracted by Tesseract OCR.}
  \label{fig:dataset}
\end{figure*}

The ubiquity of computers and smartphones has incentivized governments and companies alike to digitize most of their processes. 
Onboarding new clients, paying taxes and proving one's identity is more and more done through a computer, as the rise of online banking has shown in the last few years.
Industrial and public archives are also ongoing serious efforts to digitize their content in an effort for preservation, e.g. for old manuscripts, maps and documents with a historical value.
This means that previously physical records, such as forms and identity documents, are now digitized and transferred electronically.
In some cases, those records are produced and consumed by fully automated systems that rely on machine-readable formats, such as XML or PDF with text layers.
However, most of these digital copies are generated by end-users using whatever mean they have access to, i.e. scanners and cameras, especially from smartphones.
For this reason, human operators have remained needed to proofread the documents, extract selected fields, check the records' consistency and ensure that the appropriate files have been submitted.
Automation through expert systems and machine learning can help accelerate this process to assist and alleviate the burden of this fastidious work for human workers.

A common task involved in data filing processes is document recognition, on which depends the class-specific rules that command each file.
For example, a user might be asked to upload several documents such as a filled subscription form, an ID and a proof-of-residence.
In this work, we tackle the document classification task to check that all required files have been sent so that they are filed accordingly.

Yet, if discriminating between broad classes of documents can be achieved based on their appearance only (e.g. separating passports from banking information), fine-grained recognition often depends on the textual content of the documents.
For example, different tax forms might share their layout, logos and templates while the content in itself vastly differs.
Computer vision has been interested for some time in optical character recognition (OCR) to extract text from images.
However, dealing with both the textual and visual contents remains an open problem.
In the past years, deep learning has been established as the new state-of-the-art for image classification and natural language processing. For fine-grained document recognition, we expect the model to leverage both image and text information.

This work introduces a multimodal deep network that learns from both a document image and its textual content automatically extracted by OCR to perform its classification.
We design a pragmatic pipeline for end-to-end heterogeneous feature extraction and fusion under time and cost constraints.
We show that taking both the text and the document appearance into account improves both single modality baselines by several percents on two datasets from the document recognition literature.
We detail some limitations of the current academic datasets and give leads for an application in an industrial setting with unclean data, such as photographed documents.

\section{Related work}

Analyzing digitized documents is an old task in computer vision that was boosted by the dissemination of computers in offices and then of digital cameras and smartphones in everyday life.
To allow for textual search and easy indexing, the critical part of digitization is extracting text content from documents that have been scanned or photographed.
Indeed, either when scanning or taking a picture of the document, its actual text is lost, although it is implicitly embedded in the pixel values of the image.
Numerous optical character recognition (OCR) algorithms have been designed to transform images into strings of characters~\cite{wong_1982_document,kay_2007_tesseract}. 
Despite those efforts perfectly reading any type of document remains challenging due to the wide variety of fonts and languages.
Layout analysis is a way to preprocess the data to detect text areas and find the text orientation in order to enforce a better local and global consistency~\cite{le_1995_classification,imade_1993_segmentation}.

Document image analysis is also one of the first topic where modern deep learning has been applied.
The first convolutional neural network (CNN)~\cite{lecun_1998_gradientbased} was originally designed for classification of digits and letters.
The computer vision community deployed consequent efforts to achieve image-based document classification without text, as shown by a 2007 survey~\cite{chen_2007_survey} which focuses on document image classification without OCR results.
As an example, \cite{kumar_2014_structural} introduced SURF visual features with a bag-of-words scheme to perform document image classification and retrieval.
In 2015, \cite{harley_2015_evaluation} introduced a large labeled image document dataset which sparked interest and generated several studies of deep CNN on this topic~\cite{tensmeyer_2017_analysis,afzal_2017_cutting,das_2018_document}, inspired by the success of these networks on ImageNet and tuning data augmentation policies, transfer learning strategies and domain adaptation for document classification.
In the same idea, \cite{sicre_2017_identity} also investigated such deep architectures to classify identity documents.
\cite{aresoliveira_2018_dhsegment} goes even further by trying to segment the full layout of a document image into paragraphs, titles, ornaments, images etc.
These models focus on extracting strong visual features from the images to classify the documents based on their layout, geometry, colors and shape.

On the other hand, text-based document classification has also long been investigated.
In 1963, \cite{borko_1963_automatic} introduced an algorithmic approach to classify scientific abstracts.
More recently, \cite{manevitz_2001_oneclass} experimented with one-class SVM for document classification based on various text features, such as TF-IDF.
\cite{rubin_2012_statistical} used Latent Dirichlet Allocation to perform topic modeling and used it as a generative approach to document classification.
The recent appearance of learned word embeddings approaches such as word2vec~\cite{mikolov_2013_efficient} or ELMo~\cite{peters_2018_deep} paved to way to a large body of works related to recurrent and attention mechanisms for text classification.
For example, \cite{yang_2016_hierarchical} proposed a bidirectional recurrent network with a hierarchical attention mechanism that learns both at the word and sentence levels to improve document classification.

Some works tried to reconcile the text-based and image-based approaches to exploit both information sources.
\cite{noce_2016_embedded} performs OCR to detect keywords in images which are then encoded as colored boxes before passing the image through a CNN.
While a clever trick, this does not leverage the representation power of word embeddings.
Closer to our approach, \cite{yang_2017_learning} goes further by generating text feature maps that are combined with visual feature maps in a fully convolutional network.
However, the considered documents are synthetic and the network is trained using perfectly clean texts and images, which is unrealistic for practical uses.
More similar to us, \cite{augereau_2014_improving} learns to combine bag of words and bag of visual words features for industrial document images using a statistical model combining outputs of two single-modality classifiers.
While using shallow features, they show that using both information allows for a better accuracy when the OCR is unreliable, which is often the case in an industrial setting.

\begin{figure}[t]
  \begin{subfigure}{0.48\linewidth}
    \includegraphics[width=\textwidth]{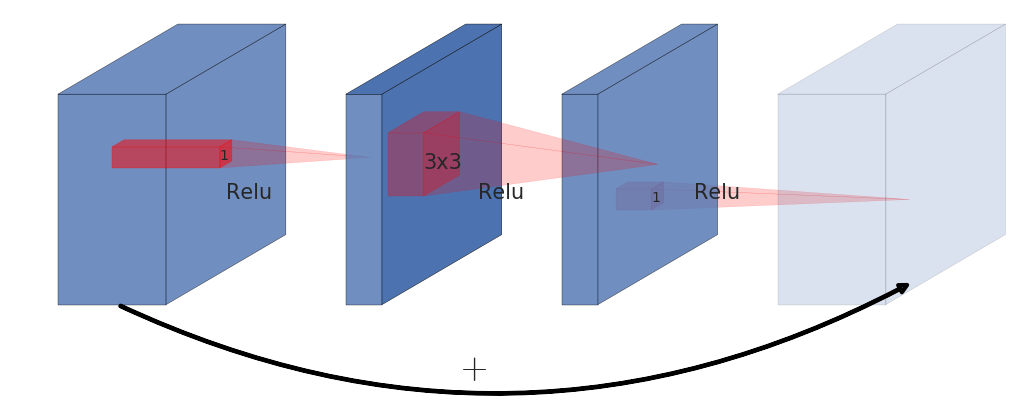}
    \caption{Residual block}
  \end{subfigure}%
  \begin{subfigure}{0.48\linewidth}
    \includegraphics[width=\textwidth]{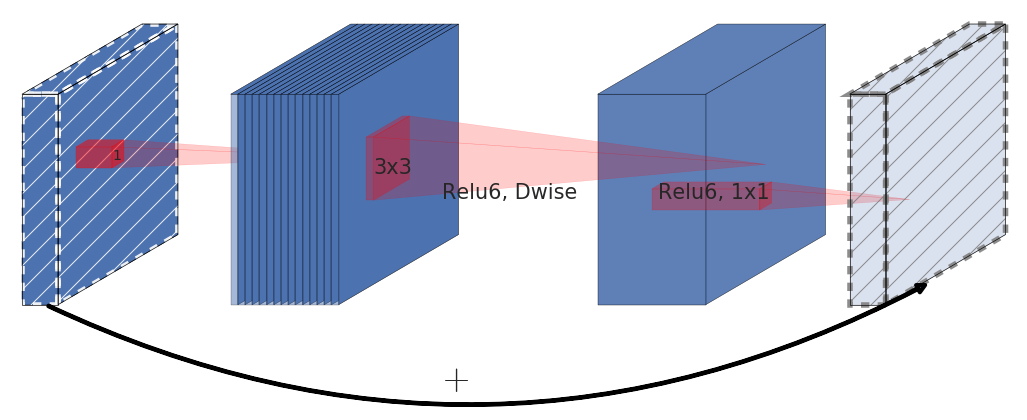}
    \caption{Inverted residual block}
  \end{subfigure}
  \caption{MobileNetV2 uses inverted residual blocks to reduce the number of channels that are forwarded in subsequent layers. Figure from~\cite{sandler_2018_mobilenetv2}.}
  \label{fig:inverted_residual}
  \end{figure}

In this paper, we go further in this direction and propose a new baseline with a hybrid deep model. In order to classify OCRized document images, we present a pragmatic pipeline perform visual and textual feature extraction using off-the-shelf architectures. To leverage the complementary information present in both modalities, we design an efficient end-to-end network that jointly learn from text and image while keeping computation cost at its minimum. We build on existing deep models (MobileNet and FastText) and demonstrate significant improvements using our fusion strategy on two document images dataset.

\section{Learning on text and image}

\subsection{Visual features}

There is a large literature both in general image recognition and in image document classification.
Recent works have established deep convolutional neural networks as the \emph{de facto} state of the art on many competitions in object recognition, detection and segmentation, e.g. ImageNet.
Deep features, extracted by pretrained or fine-tuned deep CNNs, constitute a strong baseline for visual recognition tasks~\cite{razavian_2014_cnn}.
Based on this, we choose to fine-tune a CNN pretrained on ImageNet in order to extract visual features on our images, as suggested in several recent document classification publications~\cite{tensmeyer_2017_analysis,afzal_2017_cutting,harley_2015_evaluation}
As we aim to perform inference on a large volume of data with time and cost constraints, we focus on a lightweight architecture with competitive classification performance, in our case the MobileNet v2 model~\cite{sandler_2018_mobilenetv2}.

MobileNetV2~\cite{sandler_2018_mobilenetv2} consists in a stack of bottleneck blocks.
Based on the residual learning principle~\cite{he_2016_deep}, each bottleneck block transforms a feature map first by expanding it by increasing its number of channels with a \num{1x1} convolutional layer with identity activation.
Then, a \num{3x3} depthwise convolution is performed, followed by a ReLU and a final \num{1x1} convolution with ReLU.
For efficiency issues, this block inverts the traditional residual block since the expansion is performed inside the block, whereas residual blocks compress and then reexpand the information, as illustrated in~\cref{fig:inverted_residual}.
The final MobileNetV2 contains 19 residual bottleneck layers.
Compared to other state of the art CNNs, MobileNetV2's accuracy is on-par with VGG-16 while being significantly faster.



\subsection{Textual features}

Since our use case focuses on document images in which the text has not been transcribed, we need to perform an OCR step.
To this end, we use the Tesseract OCR engine~\cite{kay_2007_tesseract} in its 4.0 version which is based on an LSTM network.
Tesseract is configured in English to use full page segmentation and the LSTM engine.
In practice, this means that Tesseract will try to detect the text orientation in the image and perform the needed affine transformation and rotation if any.
Tesseract also deals with the image binarization using Otsu's thresholding to identify black text on white background~\cite{otsu_1979_threshold}.
This will suffice on the datasets described in~\cref{sec:datasets}, although we found Tesseract challenging to apply on real-world images, especially pictures which are not flat and grayscale scans.

%


Recent literature in NLP suggests that pretrained word embeddings offer a strong baseline which surpasses traditional shallow learning approaches.
Many word embeddings have been designed following the initial success of \emph{word2vec}~\cite{mikolov_2013_efficient}, such as GloVe~\cite{pennington_2014_glove} or more recently the contextualized word embeddings from ELMo~\cite{peters_2018_deep}.

However, those word embeddings assume a good tokenization of the words, i.e. most embeddings remove digits, ignore punctuation and do not deal with out-of-vocabulary (OOV) words.
Since these embeddings are learned on clean corpus (e.g. Wikipedia or novels), tokenization is fairly straightforward.
OOV words are either assigned a random embedding or mapped to the closest in-vocabulary word based on the Levenshtein distance.

\begin{figure*}[t]
    \begin{subfigure}[b]{0.35\linewidth}
        \includegraphics[width=\linewidth]{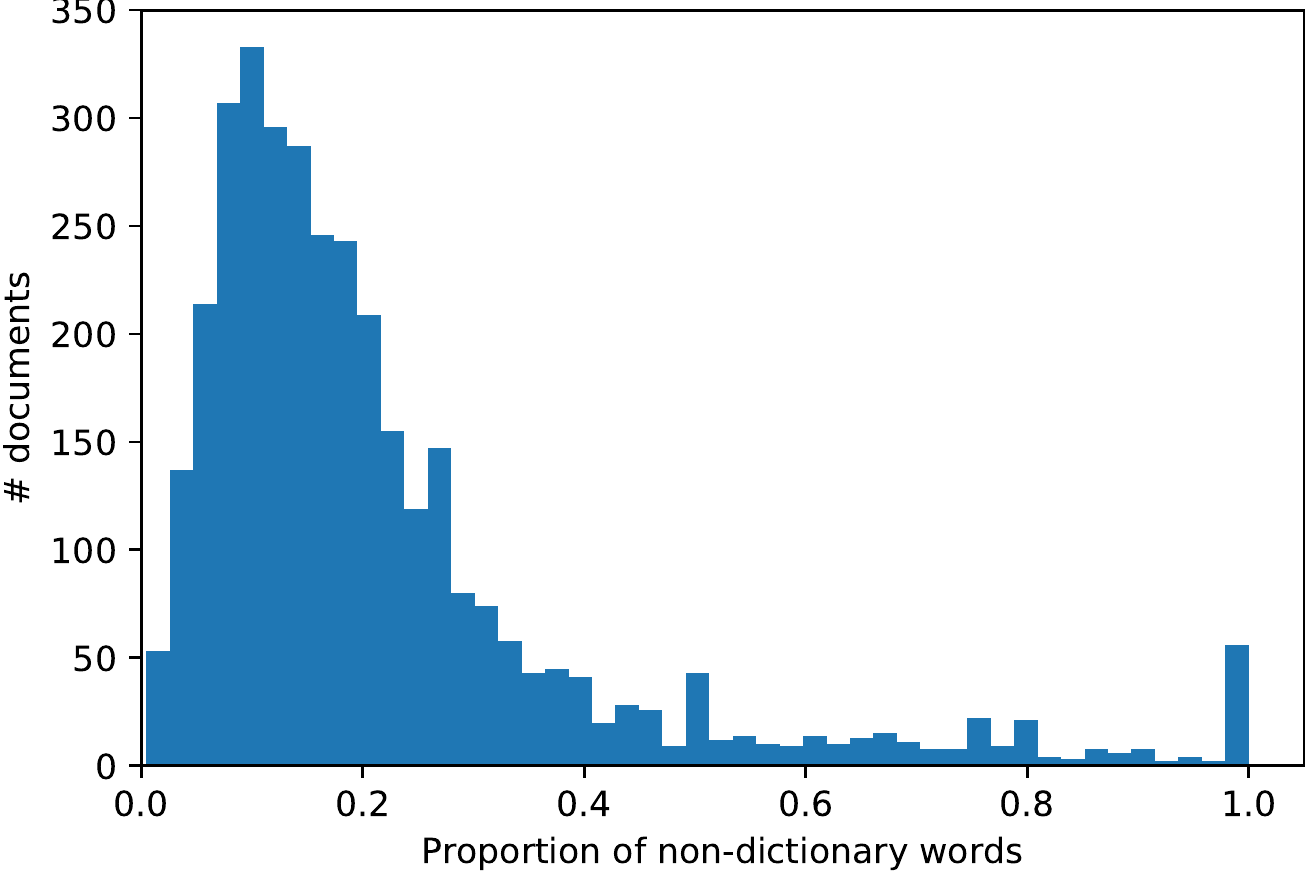}
        \caption{Document distribution w.r.t of non-dictionary words \% in the Tobacco3482-Tesseract corpus.}
        \label{fig:word_stats}
    \end{subfigure}
    \hfill
    \begin{subfigure}[b]{0.59\linewidth}
        \begin{tabularx}{\textwidth}{l Y Y Y}
        \toprule
        Word pair              & \multicolumn{3}{c}{Similarities}\\
        \midrule
        \texttt{specifically}  & GloVe & ELMo & FastText\\
        \texttt{Specificalily} & 0.71 & 0.68 & 0.96\\
        \texttt{filter}        & GloVe & ELMo & FastText\\
        \texttt{fiilter}       &  0.91 & 0.73 & 0.96\\
        \texttt{alcohol}       & GloVe & ELMo & FastText\\
        \texttt{Aleohol}       & 0.40  & 0.69 &  0.88\\
        \texttt{Largely}       & GloVe & ELMo & FastText\\
        \texttt{Largly}        & 0.25 & 0.81 &  0.98\\
        \bottomrule
        \end{tabularx}
        \vfill
        \caption{Word embeddings similarity for misspelled words.}
        \label{fig:word_embeddings}
    \end{subfigure}
    \caption{Tesseract OCR outputs noisy text that does not entirely overlap with the assumptions usually held when training word embeddings for NLP.}
\end{figure*}

Unfortunately, outputs of the Tesseract OCR are noisy and not as clean as the training data from these embeddings. Even in grayscale, well-oriented documents, OCR might have trouble dealing with diacritics, exotic fonts or curved text, as illustrated by the extracts from~\cref{fig:dataset}.
Moreover, specific user domains (e.g. banking or medieval manuscripts) might use rare words, codes, abbreviations or overall jargon that is absent from general-purpose word embeddings.
Since we face many possible misspellings in the extracted text, we cannot use the previous workarounds for OOV embeddings since it would inject a lot of non-discriminant features in our text representation
In average, on the Tobacco3482 corpus, a document processed by Tesseract OCR contains 136 words with 4 characters or more.
Of those, only 118 in average are in the GloVe embeddings~\cite{pennington_2014_glove}\footnote{Based on the Wikipedia 2014 + Gigaword 5 datasets.} and only 114 are in Enchant's spellchecker US English dictionary.
Overall, approximately 26\% of the corpus is absent from the US English dictionary and 23\% from the GloVe embeddings.
The document distribution with respect to the proportion of out-of-vocabulary words is shown in~\cref{fig:word_stats}.
Although most of the documents are concentrated around 10\% of OOVs, there is a significant long tail including several dozens of documents that contain only words outside of the English language.

Therefore, we turn to character-based word embeddings that are able to deal with OOV words by assigning them plausible word vectors that preserve both a semantic and a spelling similarity.
One possibility was to use the mimicking networks from \cite{pinter_2017_mimicking} that learn to infer word embeddings such as GloVe, but based only on subword information.
More complex embeddings such as FastText~\cite{bojanowski_2017_enriching,joulin_2017_bag} and ELMo~\cite{peters_2018_deep}, which produce vectors using respectively n-grams and subword information, can also address this problem.
Finally, the Magnitude library~\cite{patel_2018_magnitude} uses two alternative strategies to deal with OOV words:
\begin{itemize}
  \item Assigning a \emph{deterministic} random vector.
  These vectors do not capture semantic sense, however similar words based on the Levenshtein-Damerau distance will have similar vectors.
  Misspellings will therefore not be close to the original word, but similar lingo words will be close.
  \item Using character n-grams inspired by~\cite{bojanowski_2017_enriching} and interpolation with in-vocabulary words, Magnitude can generate vectors for OOV words which are sensible based on existing learned embedding.
\end{itemize}

Preliminary data exploration shows that subword-aware embeddings perform better at preserving similarity despite misspellings, as illustrated in~\cref{fig:word_embeddings}.
We therefore focus our interest on the FastText embedding, which is faster than ELMo since the latter requires passing the context through a bidirectionnal LSTM during inference.
It is worth noting that this raises concern for characters that have not been seen by FastText.
We found experimentally that Tesseract OCR generated no character that was OOV for FastText on the documents we considered.

Finally, it is necessary to convert those word embeddings into a document embedding.
We consider two approaches:
\begin{itemize}
    \item The simple baseline for sentence embedding suggested in \cite{arora_2016_simple}, which consists in a weighted average of word embeddings altered by PCA.
    \item Using variable-length document embeddings consisting in a sequence of word embeddings.
\end{itemize}

The first approach is suitable as generic feature while the second requires a statistical model able to deal with sequences, such as recurrent or convolutional neural networks.
For both methods, we use the SpaCy small English model~\cite{honnibal_2017_spacy} to perform the tokenization and punctuation removal.
Individual word embeddings are then inferred using FastText~\cite{bojanowski_2017_enriching} pretrained on the Common Crawl dataset.

\subsection{Multimodal features}

Once text and image features have been extracted, we feed them to a multi-layer perceptron following~\cite{eitel_2015_multimodala}.
To do so, we need to combine both feature vectors into one.
Two approaches can be envisioned:
\begin{itemize}
  \item Adaptive averaging of both feature vectors.
  This aligns both feature spaces so that scalars at the same index become compatible by summation, i.e. that each dimension of the vectors have a similar semantic meaning.
  \item Concatenating both vectors.
  This does not imply that both feature spaces can be aligned and delegates to the fusion MLP the task of combining the two domains.
\end{itemize}

Both fusion strategies are differentiable, therefore the whole network can be trained in an end-to-end fashion.
Moreover, the model is modular and each feature extractor can be swapped for another model, e.g. MobileNet can be exchanged with any other popular CNN and FastText could be replaced by subword-level NLP models, even differentiable ones that could allow fine-tuning the embeddings.
In this work, we try to keep things simple and build on robust base networks in order to clearly understand how the data fusion impacts model performance.
Preliminary experiments showed that the summation fusion significantly underperformed compared to pure image baseline.
We suggest that this is provoked by the impossibility of aligning the text and image feature spaces without breaking their discriminating power, resulting in suboptimal space.
Therefore, we move on with the concatenation strategy for the rest of this paper.
The complete pipeline is illustrated in~\cref{fig:multimodal_network}.

\section{Experimental setup}

\subsection{Datasets}
\label{sec:datasets}

\subsubsection{Tobacco3482}
The Tobacco3482 dataset~\cite{kumar_2014_structural} contains \num{3482} black and white documents, a subset from the Truth Tobacco Industry Documents\footnote{\url{https://www.industrydocuments.ucsf.edu/tobacco/}} archives of legal proceedings against large American tobacco companies.
There are annotations for 10 classes of documents (e.g. email, letter, memo\dots). Following common practices, we perform k-fold cross-validation using 800 documents for training and the rest for testing. Results are averaged over 3 runs.

\subsubsection{RVL-CDIP}
The RVL-CDIP dataset~\cite{harley_2015_evaluation} is comprised of \num{400000} grayscale digitized documents from the Truth Tobacco Industry Documents.
There are annotations for 16 classes of documents (e.g. email, letter, invoice, scientific report\dots), each containing \num{25000} samples. We use the standard train/val/test split from~\cite{harley_2015_evaluation} with \num{320000} documents for training, \num{40000} for validation and \num{40000} for testing.

\subsubsection{Text generation}
The Tobacco3482 and RVL-CDIP are image-based datasets. In order to evaluate our multi-modal networks, we wish to learn from both visual and textual content.
Therefore we use the Tesseract OCR library\footnote{\url{https://github.com/tesseract-ocr/tesseract/}} to extract text from the grayscales images.
We perform this operation on both datasets.
We release the OCR text dataset openly\footnote{The QS-OCR dataset is available at: \url{https://github.com/Quicksign/ocrized-text-dataset}} to encourage other researchers to replicate our work or test their own model for post-OCR text classification or multi-modal text/image classification.

\subsection{Models}

This subsection describes the implementation details of our deep networks. 
All models are implemented in TensorFlow 1.12 using the Keras API and trained using a NVIDIA Titan X GPU.
Hyperparameters were manually selected on a subset of Tobacco3482 and fixed for all experiments.

\begin{table*}[t]
    \setlength\tabcolsep{8pt}
    \begin{subfigure}[t]{0.49\linewidth}
        \caption{Preliminary experiments on Tobacco3482 for the text baseline.}
        \label{table:text_preliminary}
        \begin{tabularx}{\linewidth}{X c c}
            \toprule
            Model & OA & $F_1$\\
            \midrule
            MLP  {\scriptsize (document)} & \num{70.8}\%   & \num{0.69}\\
            CNN 1D {\scriptsize (word sequence)} & \num{73.9}\%  & \num{0.71}\\
            \bottomrule
        \end{tabularx}
        \tiny{OA = overall accuracy, $F_1$ = class-balanced $F_1$ score.}
    \end{subfigure}
    \hfill
    \begin{subfigure}[t]{0.49\linewidth}
        \caption{Preliminary experiments on Tobacco3482 for the image baseline.}
        \label{table:image_preliminary}
        \begin{tabularx}{\linewidth}{X c c}
            \toprule
            Model & OA & $F_1$\\
            \midrule
            MobileNetV2 & \num{84.5}\%  &  \num{0.82}\\
            MobileNetV2 (w/ DA) & \num{83.9}\% & \num{0.82}\\
            \bottomrule
        \end{tabularx}
        \tiny{OA = overall accuracy, $F_1$ = class-balanced $F_1$ score, DA = data augmentation.}
    \end{subfigure}
    \caption{Preliminary tuning of the single-modality baselines on Tobacco3482.}
\end{table*}

\subsubsection{Text baseline}

Seeing that our representation of textual data can be either a document embedding or a sequence of word embeddings, we compare two models for our text baseline.

The first model is an improved Multi-Layer Perceptron (MLP) with ReLU activations, Dropout and Batch Normalization (BN) after each layer.
The network has a fixed width of \num{2048} neurons for all layers except the last one, which produces a \num{128} feature vector, classified by a softmax layer.
Weights are randomly initialized using He's initialization~\cite{he_2015_delving}.
The averaged document embedding~\cite{arora_2016_simple} is used as an input for this classifier.

The second model is a one-dimensional convolutional neural network designed inspired by previous work for sentence classification~\cite{kim_2014_convolutional}.
The CNN is 4-layers deep and interlaces 1D convolutions with a window of size 12 with maxpooling with a stride of 2.
Each layer consists in 512 channels with ReLU activation.
The final feature map is processed by a max-pooling-through-time layer that extracts maximal features on the sequence on top of which we apply Dropout for regularization.
A fully connected layer then maps the features to the softmax classifier.
The input word sequence is zero-padded up to 500 words for documents with less 500 words.

We experiment on the Tobacco3482 dataset in order to evaluate which text model to choose.
Results are reported in~\cref{table:text_preliminary}.
Without surprise, the CNN 1D outperforms significantly the MLP classifier.
The pattern recognition abilities of the convolutional network makes it possible to interpret the word sequences by leveraging contextual information.
Since only some part of the text might be relevant, averaging over all word embeddings dilute the discriminating information.
Moreover, noisy embeddings due to garbage output from Tesseract (e.g. incoherent strings where OCR has failed) are included in the final document embedding.
However, when dealing with word sequences, convolutional layers and temporal max-pooling help extracting only the relevant information.
Therefore, we choose to include the 1D CNN as the text component in our multimodal architecture.

This model is denoted \tex in the rest of the paper.
It is optimized using Stochastic Gradient Descent with momentum for \num{100} epochs, with a learning rate of \num{0.01}, a momentum of \num{0.9} and a batch size of 40\footnote{Hyperparameters are manually tuned on a small validation set.}.



\subsubsection{Image baseline}

We investigate as our base CNN the lightweight MobileNetV2~\cite{sandler_2018_mobilenetv2} which focuses on computing efficiency, albeit at the cost of a slightly lower top-1 accuracy on ImageNet compared to other state of the art CNN.
We train the CNN on grayscale document images resized at \num{384x384}.
Although this warps the aspect ratio, \cite{tensmeyer_2017_analysis} reports better accuracy than when using padding at the same resolution.
As the model is designed for RGB images, the grayscale channel is duplicated three times.
This allows us to initialize the network by loading its pretrained weights on ImageNet, which accelerates convergence and slightly improves accuracy through transfer learning.

This model is denoted \vis in the rest of the paper.
It is optimized using Stochastic Gradient Descent with momentum for 200 epochs, with a learning rate of \num{0.01}, a momentum of \num{0.9} and a batch size of 40.

As reported in \cref{table:image_preliminary}, preliminary experiments on the Tobacco3482 with random JPEG artifacts, saturation and contrast alterations did not significantly alter the classifier's accuracy compared to no augmentation.
This is explained by the low variability between the grayscale document images.
All images are grayscale with dark text on white background with horizontal text lines, therefore color and geometric augmentation are not necessary.
However, \cite{tensmeyer_2017_analysis} report some success using shear transform, which we did not consider in this work.
It is worth noting that compared with previous literature on the RVL-CDIP dataset, e.g.~\cite{tensmeyer_2017_analysis,afzal_2017_cutting,harley_2015_evaluation}, we do not average predictions over multiple crops at inference time for speed concerns.
This might explain why our visual baseline underperforms the current state of the art in this state (although this does not question the gains due to the multi-modal network).

\begin{table*}
  \caption{Overall accuracy on the RVL-CDIP dataset.}
  \label{table:rvl_results}
  \setlength\tabcolsep{4pt}
  \begin{tabularx}{\linewidth}{r Y Y Y Y Y Y}
  \toprule
  Model     & \vis    & \tex   & \fus   & CNNs~\cite{harley_2015_evaluation} & VGG-16~\cite{afzal_2017_cutting} & AlexNet+SPP~\cite{tensmeyer_2017_analysis}\\
  \midrule
  OA        & 89.1\%  & 74.6\% & 90.6\% & 89.8\% & 90.97\% & 90.94\%\\
  \bottomrule
  \end{tabularx}
  \footnotesize{OA = Overall Accuray.}
\end{table*}

\begin{table*}
    \setlength\tabcolsep{3pt}
    \caption{Overall accuracy and $F_1$ scores on the Tobacco3482 datasets.}
    \label{table:tobacco_results}
    \begin{tabularx}{\linewidth}{r c Y Y Y Y Y Y Y Y Y Y Y}
        \toprule
        Model & OA & $F_1$ & Adv. & Email & Form & Letter & Memo & News & Notes & Report & Res. & Sci.\\
        \midrule
        CNNs~\cite{harley_2015_evaluation} & 79.9 & -- & \multicolumn{10}{c}{--}\\
        \midrule
        \tex  & 73.8 & 0.71 & 0.60 & 0.96 & 0.76 & 0.71 & 0.79 & 0.67 & 0.62 & 0.43 & \textbf{0.97} & 0.57\\
        \vis  & 84.5 & 0.82 & \textbf{0.94} & 0.96 & 0.85 & 0.83 & 0.90 & 0.89 & 0.83 & 0.61 & 0.80 & 0.62\\
        \fus  & \textbf{87.8} & \textbf{0.86} & 0.93 & \textbf{0.98} & \textbf{0.88} & \textbf{0.86} & \textbf{0.90} & \textbf{0.90} & \textbf{0.85} & \textbf{0.71} & 0.96 & \textbf{0.68}\\
        \midrule
        Oracle& 92.1 & 0.91 & 0.94 & 0.99 & 0.94 & 0.92 & 0.93 & 0.93 & 0.89 & 0.81 & 0.97 & 0.79\\
        \bottomrule
    \end{tabularx}
    {\scriptsize Adv. = Advertisement, Res. = Resume, Sci. = Scientific.}
\end{table*}




\subsubsection{Fusion}

For our multimodal network, we consider the same model as our baselines except that the final layers are cut-off.
For the \tex model, the last layer produces an output vector of dimension \num{128} instead of the number of classes.
For the \vis model, we aggregate the last convolutional features using global average pooling on each channel, which produces a feature vector of dimension \num{1280}.
We then map this feature vector using a fully connected layer to a representation space of dimension \num{128}.

This model is denoted \fus in the rest of the paper.
It is optimized using Stochastic Gradient Descent with momentum for 200 epochs, with a learning rate of \num{0.01}, a momentum of \num{0.9} and a batch size of 40.





\section{Discussion}

\subsection{Performances}

Model performances scores on Tobacco3482 and RVL-CDIP are reported in \cref{table:tobacco_results,table:rvl_results}.
Behaviour of all models is consistent both on the smaller dataset and on the very large one.
In both cases, the \tex baseline is significantly underperforming the \vis one.
Indeed, as could be seen in~\cref{fig:dataset}, Tesseract OCR outputs noisy text.
This includes words that have been misspelled -- which are correctly dealt with by the FastText embeddings -- and new words that are hallucinated due to poor binarization or salt-and-pepper noise in the image.
Moreover, layout and visual information tends to be more informative based on how the classes were defined: scientific papers, news and emails follow similar templates while advertisements present specific graphics.
However, in both cases, this simple document embedding is enough to classify more than 70\% of the documents, despite its roughness.

Using the \vis model only, we reach accuracies competitive with the state of the art.
MobileNetV2 alone does on-par is with the holistic CNN ensemble from~\cite{harley_2015_evaluation} and is competitive with fine-tuned GoogLeNet and ResNet-50~\cite{afzal_2017_cutting} (\num{90.97}\%).

On both datasets, the fusion scheme is able to improve the overall accuracy by $\simeq$\num{1.5}\% which demonstrates the relevance of our approach.
While the document embedding we chose is simple, it appears to be at least partially robust to OCR noise and to preserve enough information about the document content to boosts CNN accuracy on document image classification even further.
We also report the results from an oracle, which corresponds to the perfect fusion of the \tex and \vis baselines, i.e. a model that would combine the predictions from both single-modality networks and always choose the right one.
The oracle corresponds to the theoretical maximal accuracy boost that we could expect from the \fus model.
On Tobacco3482, the oracle corresponds to a \num{7.6}\% absolute improvement (9\% relative).
In our case, the \fus model improves the best single-source baseline by an absolute \num{3.3}\% (4\% relative), which is significant although still leaves the door open to further improvements.
More importantly, the gains are consistent on all classes of interest, almost never underperforming one of the two base networks on any class.
This confirm the proposed approach as the two sources, image and text, give complementary information to classify a document.

\subsection{Processing time}

Although some applications of document image recognition can be performed offline, most of the time users upload a document and expect near real-time feedback.
User experience engineering~\cite{nielsen_1993_usability} indicates than less than \si{1\second} is the maximum latency the user can suffer before the interface feels sluggish, and \si{10\second} is the maximum delay before they start loosing their attention.
On the RVL-CDIP dataset, Tesseract processes a document image in $\simeq$\si{910\milli\second} in average on an Intel Core i7-8550U CPU using 4 threads, including loading the image from disk.
This means that every additionnal latency induced by the network inference time is critical since it will negatively affect the user experience.

On the same CPU, the full inference using the \fus model takes $\simeq$\si{360\milli\second} including loading, resizing and normalizing the image.
The complete process including Tesseract OCR therefore takes less than $\simeq$\si{1300\milli\second} which is acceptable in a system requiring user input.
Of those, \si{130\milli\second} are spent in the 1D CNN (including reading the file and performing FastText inference) and \si{230\milli\second} in MobileNetV2 (including image preprocessing).
The overhead added by the final fusion layer is negligible.
We stress that this is using a standard TensorFlow without any CPU-specific compilation flags, which could speed up the inference further.
On a NVIDIA Titan X GPU, the \fus network runs in \si{110\milli\second} (\si{50\milli\second} for \tex, \si{60\milli\second} for MobileNetV2), which brings the total just above the \si{1\second} recommendation.
In our case, using compute-efficient architectures allow us to avoid running on an expensive and power-hungry GPU.

As a comparison basis, other architecture choices that we dismissed earlier would have resulted in poorer performance and the network would not be usable in a near real-time user application.
For example, the Xception network~\cite{chollet_2017_xception} takes \si{630\milli\second} to run during inference with the same parameters and hardware.
For the text model, an LSTM-based RNN with a similar depth takes many seconds to run.

Note that, although this does not reduced the perceived delay for one user, the global throughput of the system can be improved by batching the images.
Two Tesseract processes can leverage the full eight cores from an Intel Core i7-8550U CPU.
In this setting, processing an image takes $\simeq$\si{660\milli\second} in average.
Thanks to the batch efficiency of neural networks, the average processing time becomes $\leq$\si{750\milli\second} on GPU and $\leq$\si{1000\milli\second} on CPU.
This is particularly helpful when users have several documents to upload that can be processed concurrently.

\subsection{Limitations}

One of the main limitation of this work stems from the public document image datasets available.
Indeed, in a real-world application, document images can be grayscale, RGB, scanned images and photographs with various rotations, brightness, contrast and hue values.
The Tobacco documents are all oriented in the right way, which makes it easier for Tesseract to perform OCR.
Moreover, documents have been scanned by professionals who tried to maximize their legibility while user-generated often presents poor quality.

While it was not required here, data augmentation is definitely required for practical applications to encompass the large variety of environmental conditions in which documents are digitized.
This is especially true for rotations, since it is often not possible to ensure that users will capture the document with the right orientation and Tesseract does not always correctly detects it.
For industrial-grade applications dealing with user-generated content, such a data augmentation is necessary to alleviate overfitting and reduce the gap between train and actual data.
Preprocessing page segmentation and layout analysis tools, such as dhSegment~\cite{aresoliveira_2018_dhsegment} can also bring significant improvements by renormalizing image orientation and cropping the document before sending it to the classifier.

Moreover, as we have seen, the post-OCR word embeddings include lots of noisy or completely wrong words that generate OOV errors.
In practical applications, we found beneficial to perform a semantic tokenization and named entity recognition using SpaCy.
This allows us to perform a partial spellchecking, e.g. using symspell~\footnote{\url{https://github.com/wolfgarbe/SymSpell}} to correct words that have been misread by Tesseract, without affecting proper nouns or domain-specific abbreviations and codes.
If this can deal frequent mispellings of words, it might also suppress out-of-vocabulary words such as alphanumeric codes.
Therefore, learning domain specific, character-based or robust-to-OCR embeddings~\cite{malykh_2018_robust} is an interesting lead for future research, as the current interest in the ICDAR2019 competition on Post-OCR Text Correction shows\footnote{\url{https://sites.google.com/view/icdar2019-postcorrectionocr}}.

\section{Conclusion}

In this work, we tackled the problem of document classification using both image and text contents.
Based only on an image of a digitized document, we try to perform a fine-grained classification using visual and textual features.
To do so, we first used Tesseract OCR to extract the text from the image.
We then compute character-based word embeddings using FastText on the noisy Tesseract output and generate a document embedding which represents our text features.
Their counterpart visual features are learned using MobileNetv2, a standard CNN from the state of the art.
Using those pragmatic approaches, we introduce an end-to-end learnable multimodal deep network that jointly learns text and image features and perform the final classification based on a fused heterogeneous representation of the document.
We validated our approach on the Tobacco3482 and RVL-CDIP datasets showing consistent gains both on small and large datasets.
This shows that there is a significant interest into hybrid image/text approach even when clean text is not available for document image classification and we aim to further investigate this topic in the future.







\small
\bibliography{QS_apia}
\bibliographystyle{ieeetr}

\end{document}